\documentclass{article}
\usepackage{float} 
\usepackage{microtype}
\usepackage{graphicx}
\usepackage{subfigure}
\usepackage{booktabs} 
\usepackage{hyperref}
\usepackage{multirow}
\usepackage{makecell}
\usepackage{tabularx} 
\usepackage{array}    
\usepackage{listings}
\usepackage{xcolor} 
\usepackage{courier} 

\lstdefinestyle{abcnotation}{
    backgroundcolor=\color{lightgray},   
    basicstyle=\fontsize{9pt}{10pt}\ttfamily, 
    breakatwhitespace=false,             
    breaklines=true,                     
    captionpos=b,                        
    keepspaces=true,                     
    showspaces=false,                    
    showstringspaces=false,              
    showtabs=false,                      
    tabsize=2,                           
    frame=single,                        
    frameround=ffff,                     
    framesep=3pt,                        
    framerule=0pt                        
}

\lstdefinestyle{cpu}{
    backgroundcolor=\color{lightgray},   
    basicstyle=\fontsize{7.5pt}{10pt}\ttfamily, 
    breakatwhitespace=false,             
    breaklines=true,                     
    captionpos=b,                        
    keepspaces=true,                     
    showspaces=false,                    
    showstringspaces=false,              
    showtabs=false,                      
    tabsize=2,                           
    frame=single,                        
    frameround=ffff,                     
    framesep=3pt,                        
    framerule=0pt                        
}

\usepackage[accepted]{icml2024}

\usepackage{amsmath}
\usepackage{amssymb}
\usepackage{mathtools}
\usepackage{amsthm}

\usepackage[capitalize,noabbrev]{cleveref}

\theoremstyle{plain}

\theoremstyle{definition}

\theoremstyle{remark}

\usepackage[textsize=tiny]{todonotes}

\icmltitlerunning{Beyond Language Models: Byte Models are Digital World Simulators}
\begin{document}

\twocolumn[
\icmltitle{Beyond Language Models: Byte Models are Digital World Simulators}


\icmlsetsymbol{equal}{*}

\begin{icmlauthorlist}
\icmlauthor{Shangda Wu}{msra,ccom}\hspace{-0.3em}{*}\hspace{1.2em}
\icmlauthor{Xu Tan}{msra}\hspace{-0.3em}{*}\hspace{1.2em}
\icmlauthor{Zili Wang}{indie}\hspace{1.2em}
\icmlauthor{Rui Wang}{msra}\hspace{1.2em}
\icmlauthor{Xiaobing Li}{ccom}\hspace{1.2em}
\icmlauthor{Maosong Sun}{ccom,tsinghua}
\end{icmlauthorlist}

\centering
{\small\url{https://byte-gpt.github.io}}
\centering
\icmlaffiliation{msra}{Microsoft Research Asia}
\icmlaffiliation{ccom}{Central Conservatory of Music, China}
\icmlaffiliation{indie}{Independent researcher}
\icmlaffiliation{tsinghua}{Tsinghua University, China}

\icmlcorrespondingauthor{Maosong Sun}{tsinghua.edu.cn}

\icmlkeywords{Machine Learning, ICML}

\vskip 0.3in
]



\printAffiliationsAndNotice{\icmlEqualContribution} 

\begin{figure*}[htbp] 
    \centering
    \includegraphics[width=\textwidth]{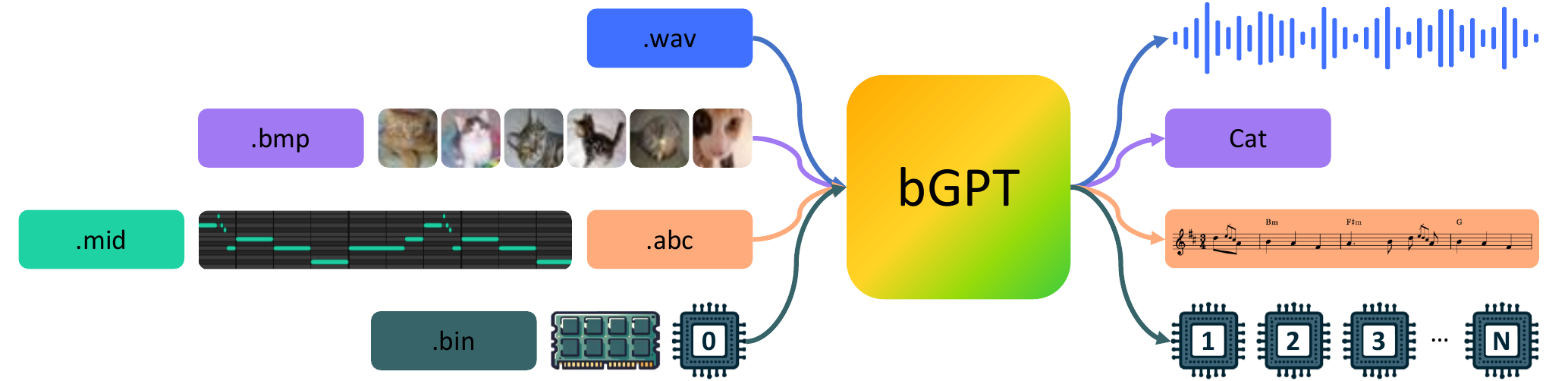} 
    \caption{The bGPT framework simulates digital systems through native binary data, and integrates diverse data types into a single model, treating everything as a byte sequence. This approach simplifies integration and expands application possibilities in the digital world.}
\vspace{-1em}
\end{figure*}

\begin{abstract}
Traditional deep learning often overlooks bytes, the basic units of the digital world, where all forms of information and operations are encoded and manipulated in binary format. Inspired by the success of next token prediction in natural language processing, we introduce bGPT, a model with next byte prediction to simulate the digital world. bGPT matches specialized models in performance across various modalities, including text, audio, and images, and offers new possibilities for predicting, simulating, and diagnosing algorithm or hardware behaviour. It has almost flawlessly replicated the process of converting symbolic music data, achieving a low error rate of 0.0011 bits per byte in converting ABC notation to MIDI format. In addition, bGPT demonstrates exceptional capabilities in simulating CPU behaviour, with an accuracy exceeding 99.99\% in executing various operations. Leveraging next byte prediction, models like bGPT can directly learn from vast binary data, effectively simulating the intricate patterns of the digital world.
\end{abstract}

\section{Introduction}
Deep learning research has focused on digital media files that are easily interpretable by humans, such as text, audio, and images \cite{oord2016wavenet,he2016deep,DBLP:conf/naacl/PetersNIGCLZ18}, due to their direct relevance to human communication and understanding. Text, in particular, plays a central role in conveying human intelligence and has led to the emergence of Language Models (LMs) \cite{radford2018improving,radford2019language,DBLP:conf/nips/BrownMRSKDNSSAA20,DBLP:journals/corr/abs-2303-12712,DBLP:journals/corr/abs-2302-13971,DBLP:journals/corr/abs-2307-09288}. The fundamental principle of LM involves tokenizing text \cite{DBLP:conf/acl/SennrichHB16a,DBLP:journals/corr/WuSCLNMKCGMKSJL16,DBLP:conf/acl/Kudo18,DBLP:conf/emnlp/KudoR18} and predicting the next token in a sequence, enabling them to comprehend human language and intelligence. Recent advancements further extend tokenization to various modalities beyond text \cite{DBLP:conf/iclr/DosovitskiyB0WZ21,DBLP:journals/corr/abs-2210-13438}, empowering LMs to attain a more holistic understanding of the real world and emulate human intelligence \cite{DBLP:journals/corr/abs-2301-02111,DBLP:journals/taslp/BorsosMVKPSRTGTZ23,DBLP:journals/corr/abs-2304-08485,DBLP:conf/icml/0008LSH23}.

These deep learning models, however, predominantly operate within the realm of media data, overlooking the omnipresent native binary data in the digital world. Bytes are the foundation of all digital data, devices, and software, from computer processors to operating systems in everyday electronics. Therefore, training models for next byte prediction can potentially lead to a paradigm shift in deep learning, allowing them to truly understand and simulate all activities in the digital world. This has practical benefits not only in conventional areas, but also in some underexplored areas such as boosting cybersecurity \cite{DBLP:conf/aaai/RaffBSBCN18}, improving computer diagnostics \cite{DBLP:conf/uss/GuoMXDS19}, optimizing data compression \cite{DBLP:journals/corr/abs-2309-10668}, and even advancing complex tasks like reverse-engineering the source code of that software from its binary representation.

In this paper, we introduce bGPT, a model designed for binary data processing and digital world modelling by next byte prediction. The digital world includes not only digital media files, traditionally the focus of deep learning models, but also extends to the intricate realm of digital systems, ranging from hardware architectures to complex algorithms. bGPT transcends traditional deep learning boundaries by directly interpreting and manipulating binary data, enabling a more intrinsic and holistic understanding of the digital world. Its advantages are two-fold: 1) \textbf{Interpreting Digital System}: By training on byte sequences, bGPT can learn the patterns of digital systems, enabling it to predict, simulate, and diagnose algorithm or hardware behaviour. This ability allows for the reconstruction of complex systems from binary data. 2) \textbf{Unified Modelling}: bGPT integrates various data types into a single framework, treating everything as a byte sequence. This simplifies modelling and allows for easy integration of various data sources.

Our experiments include two main areas: 1) well-studied tasks like generative modelling and classification on digital media data (e.g., text, audio, and images); and 2) relatively underexplored tasks intrinsic to binary-native operations, including data conversion and CPU state modelling, which represent algorithm and hardware simulation, respectively. The study of byte models not only marks a significant step towards holistic and unified deep learning, but also offers a new perspective on modelling the digital world.

\section{Background}
\subsection{Language Models}
LMs, from earlier LSTM-based \cite{DBLP:journals/neco/HochreiterS97} to recent Transformer-based models \cite{vaswani2017attention}, are crucial for both understanding and generating human language while simulating human intelligence. Text tokenization plays a fundamental role in these models, as it involves breaking down text into smaller units, such as words or subwords \cite{DBLP:conf/acl/SennrichHB16a,DBLP:journals/corr/WuSCLNMKCGMKSJL16,DBLP:conf/acl/Kudo18,DBLP:conf/emnlp/KudoR18}, which serve as the input for the model. The introduction of Generative Pre-trained Transformer (GPT) models \cite{radford2018improving,radford2019language,DBLP:conf/nips/BrownMRSKDNSSAA20} represents a significant advancement in this field. GPT models are pre-trained through self-supervised learning, particularly via next token prediction on extensive text data. This training technique, next token prediction, teaches the model to predict the most likely next token in a sequence, enabling it to capture the structure and semantics behind languages.

Next token prediction has extended its influence to various data types. In audio processing, models like AudioPaLM \cite{DBLP:journals/corr/abs-2306-12925} merge text and speech, enabling speech-to-speech translation and advanced speech recognition. MusicGen \cite{DBLP:journals/corr/abs-2306-05284} excels in conditional music generation by modelling multiple parallel streams of acoustic tokens extracted by EnCodec \cite{DBLP:journals/corr/abs-2210-13438}. In image processing, iGPT \cite{DBLP:conf/icml/ChenRC0JLS20} applies Transformer models to predict the next pixel in an image, while several vision-language models \cite{DBLP:journals/corr/abs-2304-08485,DBLP:journals/corr/abs-2304-10592,DBLP:conf/icml/0008LSH23} have emerged to bridge the gap between textual and visual data. In biochemical sequences, Tranception \cite{DBLP:conf/icml/NotinDFMGMG22} leverages autoregressive transformers and retrieval to predict protein fitness, while ProtGPT2 \cite{ferruz2022protgpt2} generates protein sequences with natural amino acid propensities. HyenaDNA \cite{DBLP:journals/corr/abs-2306-15794} extends context lengths in genomic modelling, enabling long-range sequence understanding. 

Next token prediction has empowered LMs to grasp the intricacies of human intelligence and the world. Expanding these techniques to binary data via next byte prediction could further enhance their versatility in handling digital information and simulating the digital world.

\subsection{Byte Models}
While binary data lacks the inherent structure and semantics of human-interpretable data like text, recent research efforts are exploring its modelling and information extraction, opening up new possibilities for byte models.

By modelling native binary data, systems like MalConv \cite{DBLP:conf/aaai/RaffBSBCN18} and DeepVSA \cite{DBLP:conf/uss/GuoMXDS19} have emerged as potent tools for malware detection and program analysis. MalConv employs Convolutional Neural Networks (CNNs) \cite{DBLP:conf/nips/CunBDHHHJ89} to analyze raw byte sequences in executable files, while DeepVSA enhances memory alias analysis within the context of value set analysis for postmortem program analysis. Additionally, the concept of language models compressing byte sequences \cite{DBLP:journals/corr/abs-2309-10668} introduces a novel perspective on how large pre-trained models \cite{DBLP:journals/corr/abs-2203-15556} can be utilized.

Several studies have validated the utility of byte-level encoding for language tasks. For instance, Byte-level Byte Pair Encoding (BBPE) has been used to enhance multilingual model pre-training \cite{wei2021training} and has also shown promise in machine translation \cite{DBLP:conf/aaai/WangCG20}, striking a balance between processing efficiency and linguistic breadth. ByT5 \cite{DBLP:journals/tacl/XueBCANKRR22} builds on this by using standard Transformer models for byte sequences, promoting a token-free encoding method that improves noise robustness and spelling sensitivity in multilingual scenarios.

Byte encoding has also been applied to other human-interpretable data, allowing models to work with binary representations of text, images, and diverse data types in a universal framework. For example, ByteFormer \cite{DBLP:journals/corr/abs-2306-00238} directly handles raw byte sequences converted from images and audio while maintaining versatility and privacy. MegaByte \cite{yu2023megabyte}, on the other hand, has been tested and proven to excel in modelling long byte sequences across various modalities. Inspired by MegaByte \cite{yu2023megabyte}, MambaByte \cite{wang2024mambabyte} leverages the Mamba network structure \cite{DBLP:journals/corr/abs-2312-00752} to excel in byte-level language modelling and even outperforms LMs based on subword tokenization.

Despite progress, current research often neglects native binary data, focusing on narrow tasks and overlooking the broader potential of byte models in digital world simulation. To address these issues, we employed bGPT to model native binary data and conducted comprehensive evaluations across various tasks. This approach provides a holistic assessment of byte models in various applications, offering insights into the potential of digital world modelling.

\section{Methodology}
In this section, we introduce bGPT, a model optimized for modelling digital data at the byte level. We start by presenting its hierarchical Transformer framework, which segments byte sequences into patches to manage computational efficiency. Subsequently, we present the training objectives of bGPT, including generative modelling and classification.

\subsection{Model Architecture}
Working with digital data at the byte level not only allows models to learn digital system patterns but also offers a unified approach for incorporating various data types into a single framework. However, the high granularity of bytes results in long sequences, substantially raising computational costs. This issue is more pronounced in Transformer-based models due to quadratic self-attention scaling, restricting their efficiency and scalability for handling binary data.

To address computational limitations in Transformer-based models with long byte sequences, inspired by previous work \cite{DBLP:conf/ismir/WuY0S23,yu2023megabyte,DBLP:conf/hcmir/WuLY023,yang2023uniaudio}, bGPT adapts a hierarchical Transformer architecture and segments a sequence of bytes \(B = \{b_1, b_2, ..., b_T\}\) of length \( T \) into a sequence of patches \( \mathcal{P} \), where each patch contains exactly \( S \) bytes:

\noindent
\begin{equation}
   \mathcal{P} = [P_1, P_2, \ldots, P_N],
\end{equation}
\noindent

where \( N = \lceil \frac{T}{S} \rceil \) is the number of patches, and \( P_i = [b_{(i-1)S+1}, \ldots, b_{iS}] \) for \( 1 \leq i \leq N \). If \( T \mod S \neq 0 \), the last patch \( P_N \) is defined as:

\noindent
\begin{equation}
   P_N = [b_{(N-1)S+1}, \ldots, b_T, \underbrace{e, \ldots, e}_{S - (T \mod S)} ],
\end{equation}
\noindent

where \( e \) represents the \texttt{<eop>} (end-of-patch) token used to pad the last patch \( P_N \) to the size \( S \). By segmenting byte sequences into more manageable patches, bGPT balances the need for byte-level sequence modelling with computational efficiency.

\begin{figure}[t]
    \centering
        \includegraphics[width=8.25cm]{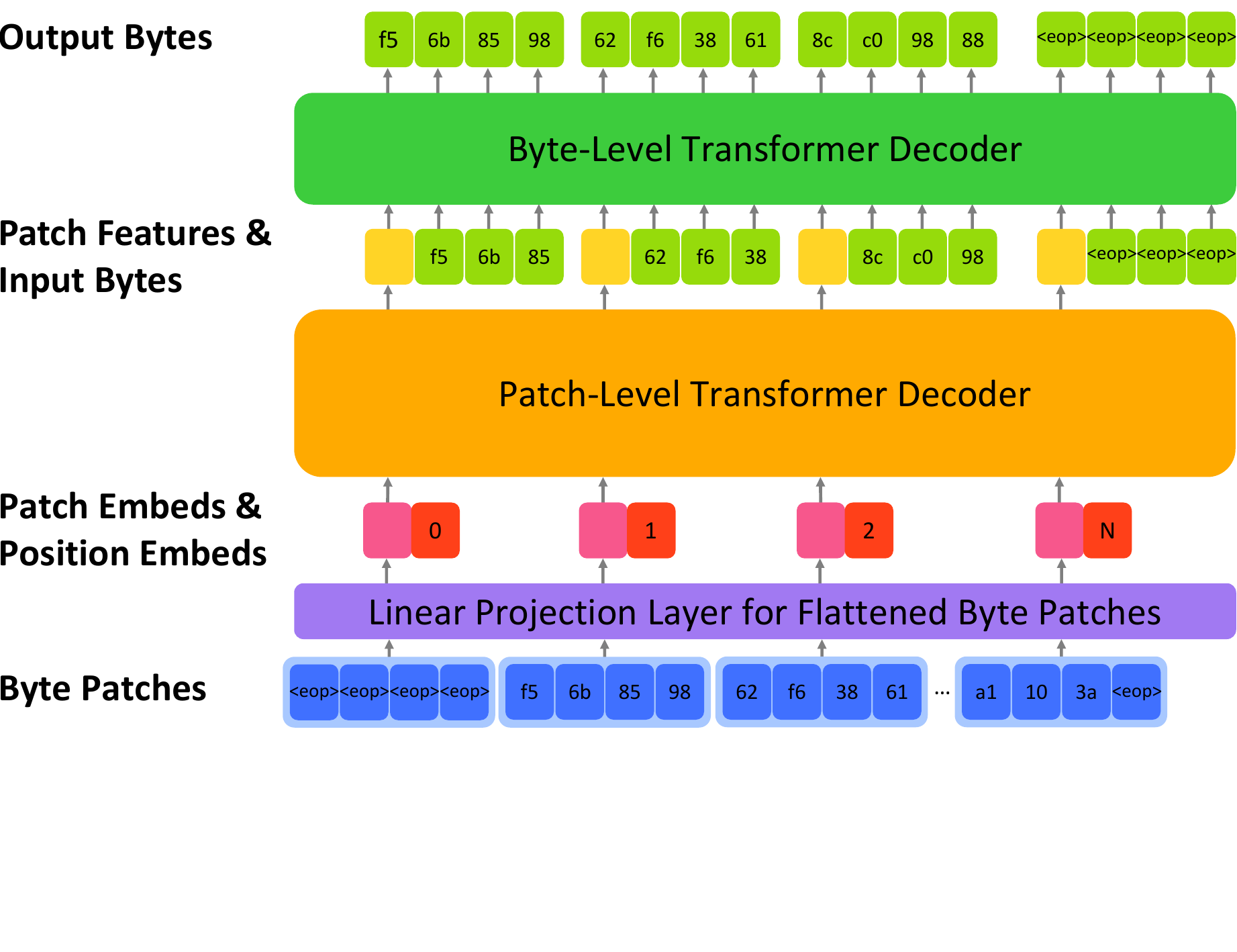}
    \centering
    \vspace{-5.5em}
	\caption{bGPT segments byte sequences into patches, predicts next patch features with a patch-level decoder, and reconstructs bytes within patches using these features with a byte-level decoder.}
    \vspace{-0.5em}
\end{figure}

As illustrated in Fig. 2, bGPT consists of three main components: a linear projection layer, a patch-level decoder, and a byte-level decoder. bGPT processes the patch sequence \( \mathcal{P} \) through its components as follows:

\textbf{Linear Projection Layer}: Each patch \( P_i \) from \( \mathcal{P} \) is viewed as a matrix of size \( S \times 257 \), where \( S \) is the patch size, and each byte is one-hot encoded into a 257D vector, including all 256 byte values and an \texttt{<eop>} token. These patches are then flattened into one-dimensional vectors, where the rows in the matrix are concatenated sequentially. The linear projection layer then maps each flattened vector into a dense vector \( E_i \)  of a hidden size \( H \). For each patch, the operation can be formulated as:

\noindent
\begin{equation}
    E_i = \text{Flatten}(P_i) \cdot W_{\text{linear}}, \quad 1 \leq i \leq N,
\end{equation}
\noindent

where \( W_{\text{linear}} \) is the weight matrix of the linear projection layer with a shape of \( (S \times 257, H) \). This dense embedding enables more efficient processing of the byte sequence by reducing the dimensionality while preserving the essential information contained in each patch.

\textbf{Patch-Level Decoder}: This decoder takes the sequence of embedded patches \( \mathcal{E} = \{E_1, E_2, \ldots, E_N\} \) and processes it to autoregressively predict the features of the subsequent patch, effectively learning the structure of the data:

\noindent
\begin{equation}
\hat{E}_{i} = \text{Decoder}_{\text{patch}}(\mathcal{E}_{< i} \oplus \mathcal{X}_{< i}),
\end{equation}
\noindent

where \( \mathcal{E}_{< i} \) denotes the sequence of patch embeddings before the \( i \)-th patch, and \( \mathcal{X}_{< i} \) represents the corresponding positional embeddings. The \( \oplus \) symbol denotes the element-wise addition of these two sequences. The output, \( \hat{E}_{i} \), is the predicted feature for the \( i \)-th patch.

\textbf{Byte-Level Decoder}: It takes the predicted feature \( \hat{E}_{i} \) of an individual patch and autoregressively reconstructs the sequence of bytes within that patch. The process is independent for each patch and operates by conditioning on the feature representation \( \hat{E}_{i} \) of the current patch:

\noindent
\begin{equation}
\hat{b}_{i, j} = \text{Decoder}_{\text{byte}}(\hat{E}_i, {b}_{i, <j}), \quad 1 \leq j \leq S,
\end{equation}
\noindent

where \( \hat{b}_{i, j} \) is the predicted byte at position \( j \) in the \( i \)-th patch, and \( {b}_{i,<j} \) represents all preceding bytes in the current patch.

\subsection{Training Objectives}
The training for bGPT primarily revolves around generative modelling through next byte prediction as its core focus, playing a pivotal role in predicting and generating bytes.

\textbf{Generative Modelling}: It aims to predict the next byte \(b_{i+1}\) in a sequence based on preceding bytes \(\{b_1, b_2, ..., b_i\}\) without explicit guidance. For a byte sequence \(B = \{b_1, b_2, ..., b_T\}\) of length \( T \), the objective is minimizing the negative log-likelihood of the next byte prediction across the sequence, defined as:

\noindent
\begin{equation}
\mathcal{L}_{\text{GEN}}(\theta) = -\sum_{i=1}^{T-1} \log p(b_{i+1} | b_1, b_2, ..., b_i; \theta),
\end{equation}
\noindent

where \(\theta\) represents the model parameters, and \(p(\cdot)\) denotes the predicted probability distribution over possible next bytes. The loss function \(\mathcal{L}_{\text{GEN}}(\cdot)\) in generative modelling encourages the model to understand the sequential dependencies in data at the byte level.

After being initially trained on generative modelling via next byte prediction, bGPT is further adapted to classification tasks with a new training objective.

\textbf{Classification}: Upon the foundation of generative modelling through next byte prediction, bGPT is further trained on labelled datasets, where it predicts categories from byte sequences. This involves extracting a global feature from the byte sequence, which is then processed by a classification head. The goal is to minimize the classification loss \(\mathcal{L}_{\text{CLF}}\), formulated as:

\noindent
\begin{equation}
\mathcal{L}_{\text{CLF}}(\theta) = -\sum_{k=1}^{K} y_k \log p(y_k | B; \theta),
\end{equation}
\noindent

where \(y_k\) is the boolean label for the \(k\)-th category, indicating whether the byte sequence belongs (true) or does not belong (false) to that category. \(K\) is the total number of categories, and \(p(y_k | B; \theta)\) is the predicted probability of category \(k\) given the byte sequence \(B\).

\section{Applications}
Byte models such as bGPT excel at understanding binary data, and have proficiency in digital media file processing (e.g., text, audio, and images) and simulating algorithms and hardware operations for in-depth software and digital system analysis. This section introduces the selected applications and corresponding experimental settings for evaluating the proposed bGPT.

\subsection{Digital Media Processing}
The field of deep learning is steadily advancing its proficiency in both the generation and classification of diverse forms of media, including text, audio, and images \cite{devlin2018bert,DBLP:conf/acl/AoWZ0RW0KLZWQ0W22,DBLP:journals/corr/abs-2204-06125}, which are essential for human communication and information exchange. These media files, typically stored and transmitted as byte sequences on electronic devices, enable bGPT to process such digital content for generative modelling and classification.

bGPT trains in generative modelling for representation learning via next token prediction. It then uses features from the final patch-level decoder layer, employing average pooling to derive global features for classification.

To streamline the training process, we standardized audio and image datasets. Audio files were converted to WAV format, with specifications including an 8000 Hz sampling rate, mono channel, and 8-bit depth, each trimmed to one-second lengths. Image data were set to BMP format with a resolution of 32$\times$32, RGB colour, and 24-bit depth.

\subsection{Algorithm and Hardware Simulation}
To demonstrate the capabilities of byte models in predicting and modelling digital processes, we select two examples—data conversion and CPU state modelling.

\textbf{Data Conversion}: The process involves converting data from one format to another, with symbolic music formats such as ABC notation and MIDI files serving as our main examples. For background information on ABC notation and MIDI, please refer to Appendix A. In this task, bGPT employs the generative modelling approach on concatenated byte sequences of paired ABC and MIDI files, separated by a special patch. The bGPT model learns to convert text-based ABC notation music scores into binary MIDI performance signals and, reversely, convert MIDI back into ABC notation. This necessitates the ability to simulate and reverse-engineer the conversion algorithm\footnote{\url{https://github.com/xlvector/abcmidi}}, which indicates an essential capability for modelling the digital world.

\begin{table*}[t!]
  \caption{Overview of datasets for bGPT evaluation, with computational costs benchmarked in NVIDIA V100 GPU hours. Details include file counts in millions, dataset sizes in gigabytes, average file sizes in bytes, and the GPU hours required per epoch for training.}
  \vspace{1em}
  \centering
  \resizebox{\linewidth}{!}{%
  \begin{tabular}{lccccc}
      \toprule
      \textbf{Dataset} & \textbf{Total Files (Million)} & \textbf{Total Bytes (GB)} & \textbf{Mean File Size (Byte)} & \textbf{GPU hours per Epoch} & \textbf{Modality} \\
      \midrule
      Wikipedia \cite{wikidump}       & 6.28                          & 13.67                     & 2237                          & 107                         & \includegraphics[height=1em]{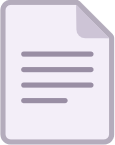}  \\
      AG News \cite{DBLP:conf/nips/ZhangZL15}        & 0.13                          & 0.03                      & 236.4                         & 2.1                         & \includegraphics[height=1em]{text.png}          \\
      ImageNet \cite{DBLP:conf/cvpr/DengDSLL009}         & 1.33                          & 3.88                      & 3126                          & 18                          & \includegraphics[height=1em]{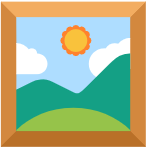}         \\
      CIFAR-10 \cite{krizhevsky2009learning}       & 0.06                          & 0.17                      & 3126                          & 1.1                         & \includegraphics[height=1em]{image.png}          \\
      LibriSpeech \cite{DBLP:conf/icassp/PanayotovCPK15}     & 3.68                          & 27.57                     & 8044                          & 84                          & \includegraphics[height=1em]{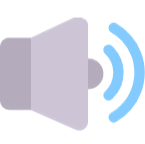}         \\
      Speech Commands v2 \cite{DBLP:journals/corr/abs-1804-03209} & 0.09                          & 0.68                      & 8044                          & 2.4                         & \includegraphics[height=1em]{speech.png}          \\
      IrishMAN-ABC \cite{DBLP:conf/hcmir/WuLY023}     & 0.21                          & 0.06                      & 313.9                         & 2.4                         & \includegraphics[height=1em]{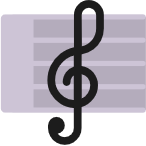} \\
      IrishMAN-MIDI \cite{DBLP:conf/hcmir/WuLY023}    & 0.21                          & 0.39                      & 1916                          & 3.2                         & \includegraphics[height=1em]{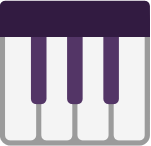}          \\
      CPU States (proposed)       & 2.1                           & 6.09                      & 3103                          & 36                          & \includegraphics[height=1em]{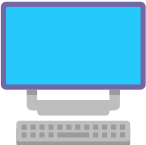}     \\
      \bottomrule
  \end{tabular}%
  }
\vspace{-1em}
\end{table*}

\textbf{CPU State Modelling}: Similarly, the model is fed with concatenated sequences of low-level machine instructions followed by a series of CPU register states. The objective is to accurately predict how the state updates with each instruction until the program halts. This task demonstrates the capacity of bGPT to interpret operational data and replicate digital activities within hardware.

For CPU state modelling, we introduce the CPU States dataset (with 2.1 million instances), offering a simplified representation of CPU behaviour for ease of data collection and evaluation. Each dataset instance contains a 1KB memory block with varying numbers of machine instructions, followed by a sequence of 16-byte CPU register states. These states include various instructions, totaling 21 unique types with 43 variants, such as data movement, logical operations, and arithmetic operations. Within each state, 1 byte each is allocated for the Program Counter (PC) and Accumulator (ACC), 4 bytes are allocated for the Instruction Register (IR), with an additional 10 bytes reserved for general-purpose registers. Instances are generated by executing random sequences of 1 to 256 instructions and capturing the state after each execution. Despite simplifications, this dataset effectively simulates typical CPU behaviour. See Appendix B for more details.

\section{Experiments}
\subsection{Settings}
Our experiments utilized open-source datasets across various domains, including images, speech, text, two types of symbolic music, and CPU states, as detailed in Table 1. The training settings were consistent across all bGPT models and datasets as per Table 2, supporting file sizes up to 8KB using a patch size of 16 and a sequence length of 512. The 110M-parameter bGPT model matches the standard Transformer-based model scale. Notably, we avoided hyperparameter tuning and data augmentation in all evaluations. We employed accuracy (Acc) as the metric to assess classification performance and Bits-Per-Byte (BPB) for other generative modelling tasks, unless otherwise specified.

\begin{table}[t]
\vspace{-0.65em}
\centering
\caption{Hyperparameter settings for pre-training and fine-tuning.}
\vspace{1em}
\resizebox{0.8\linewidth}{!}{%
    \begin{tabular}{lcc}
    \toprule
    \textbf{Hyperparameter} & \textbf{Pre-training} & \textbf{Fine-tuning} \\
    \midrule
    Patch Size        & 16 & 16 \\
    Patch Length      & 512 & 512 \\
    Patch-Level Layers & 12 & 12 \\
    Byte-Level Layers & 3 & 3 \\
    Hidden Size       & 768 & 768 \\
    Epochs            & 32 & 32 \\
    Learning Rate     & 1e-04 & 1e-05 \\
    Batch Size        & 16 & 1 \\
    \bottomrule
    \end{tabular}%
}
\vspace{-2em}
\end{table}

\begin{table*}[t!]
\centering
\caption{Performance comparison of bGPT models pre-trained on different datasets and baseline models in their respective modalities: GPT2-small on AG News, ViT-B/16 on CIFAR-10, and AST on Speech Commands v2. $\rm bGPT_{signal}$ is pre-trained on a merged dataset of ImageNet and LibriSpeech, while $\rm bGPT_{mix}$ is pre-trained on a combination of ImageNet, LibriSpeech, and Wikipedia datasets.}
\vspace{1em}
\begin{tabularx}{\textwidth}{m{2.5cm} *{6}{>{\centering\arraybackslash}X}}
\toprule
& \multicolumn{2}{c}{\textbf{AG News (4 classes)}} & \multicolumn{2}{c}{\textbf{CIFAR-10 (10 classes)}} & \multicolumn{2}{c}{\textbf{Speech Commands v2 (36 classes)}} \\
\cmidrule(lr){2-3} \cmidrule(lr){4-5} \cmidrule(lr){6-7}
\textbf{Model} & \textbf{BPB} & \textbf{Acc (\%)} & \textbf{BPB} & \textbf{Acc (\%)} & \textbf{BPB} & \textbf{Acc (\%)} \\
\midrule
$\rm bGPT_{random}$ & 1.3496 & 84.74 & 3.4928 & 76.73 & 1.5414 & 92.43 \\
$\rm bGPT_{wiki}$ & \textbf{1.0639} & \textbf{92.49} & 3.6663 & 77.02 & 1.5719 & 93.56 \\
$\rm bGPT_{image}$ & 1.4179 & 83.16 & \textbf{3.1234} & \textbf{88.69} & 1.5326 & 93.91 \\
$\rm bGPT_{libri}$ & 1.3993 & 83.59 & 3.3345 & 83.51 & \textbf{1.4818} & \textbf{96.03} \\
$\rm bGPT_{signal}$ & 1.4058 & 83.80 & 3.1554 & 87.65 & 1.4898 & 95.66 \\
$\rm bGPT_{mix}$ & 1.0935 & 91.75 & 3.2279 & 84.32 & 1.5086 & 95.09 \\
\midrule
Baselines & 0.9237 & 94.50 & —— & 98.13 & —— & 98.11 \\
\bottomrule
\end{tabularx}
\vspace{-1em}
\end{table*}

\subsection{Digital Media Processing}
The study aimed to evaluate the effectiveness of bGPT in processing digital media files at the byte level compared to specialized models. We followed the standard pre-training and fine-tuning approach in deep learning, pre-trained bGPT on diverse datasets like ImageNet ($\rm bGPT_{image}$), Wikipedia ($\rm bGPT_{wiki}$), and LibriSpeech ($\rm bGPT_{libri}$). We also explored the impact of mixed pre-training: $\rm bGPT_{signal}$ combined the ImageNet and LibriSpeech datasets, while $\rm bGPT_{mix}$ integrated all three datasets. A random-initialized variant $\rm bGPT_{random}$ was used as a baseline. These models were first fine-tuned using next byte prediction on AG News, CIFAR-10, and Speech Commands v2, and then further fine-tuned for classification.

\subsubsection{Baselines}
For baseline comparisons, we selected Transformer-based models of similar scale that excel in their respective domains: GPT2-small \cite{radford2019language} for text generation/classification, and ViT-B/16 \cite{DBLP:conf/iclr/DosovitskiyB0WZ21} and AST \cite{DBLP:conf/interspeech/GongCG21} for image and audio classification, respectively. GPT2-small was pre-trained on English Wikipedia under the same settings as bGPT. ViT and AST were pre-trained on ImageNet \cite{DBLP:conf/cvpr/DengDSLL009}, and their results were taken from the original studies. In the case of CIFAR-10 \cite{krizhevsky2009learning} and Speech Commands v2 \cite{DBLP:journals/corr/abs-1804-03209}, where generative modelling benchmarks were lacking, we did not report BPB metrics.

\subsubsection{Results}
Table 3 presents the results of various bGPT and baseline models on different benchmarks. The primary insight from this comparison is the significant influence of pre-training on model performance across all tasks. For example, $\rm bGPT_{wiki}$ pre-trained on Wikipedia performs well in text-related tasks, while $\rm bGPT_{libri}$, pre-trained on LibriSpeech, excels in spoken content tasks compared to other variants. This indicates that bGPT models can achieve better performance in downstream tasks when there is a close match between pre-training and fine-tuning datasets, aligning with standard pre-training principles \cite{DBLP:conf/nips/LiuXX00JC022}.

Despite not having modality-specific prior knowledge, bGPT models still manage to achieve performances that are on par with the baseline models. For instance, $\rm bGPT_{wiki}$ achieved a score of 1.0639 BPB and an accuracy of 92.49\% on AG News, which, while not quite at the level of GPT2-small, scored 0.9237 BPB and 94.50\% accuracy, still demonstrates competitive performance. Likewise, $\rm bGPT_{libri}$ reached an accuracy of 96.03\% on Speech Commands v2, coming close to the accuracy of AST at 98.11\%. However, in the analysis of CIFAR-10, a noticeable difference is seen with $\rm bGPT_{image}$ lagging ViT, with an accuracy of 88.69\% against 98.13\%. This discrepancy in image tasks is likely due to the sequential processing nature of byte models, which struggle to capture the essential two-dimensional spatial relationships within images \cite{yu2023megabyte}. Despite this, simply scaling the model size while retaining this sequential processing approach could still hold promise for achieving state-of-the-art results \cite{DBLP:conf/icml/ChenRC0JLS20}.

bGPT models pre-trained on mixed modalities, namely $\rm bGPT_{signal}$ and $\rm bGPT_{mix}$, yield performances that generally align with the average performance of models pre-trained on individual modalities. For example, $\rm bGPT_{mix}$ on the AG News dataset, with a BPB of 1.0935 and accuracy of 91.75\%, outperforms other variants but falls short of the performance demonstrated by $\rm bGPT_{wiki}$, which is specifically tuned for text data. Similar trends can be observed in the case of $\rm bGPT_{signal}$, which typically shows performance levels that lie somewhere between $\rm bGPT_{image}$ and $\rm bGPT_{libri}$. This illustrates a trade-off in byte models—while mixed-modality pre-training fosters versatility, it may dilute the depth of domain-specific understanding.

Another noteworthy observation from Table 3 is the mixed results in cross-modal fine-tuning on bGPT models, with both positive and negative transfer effects. Positive transfer occurs when models pre-trained on one data type (e.g., $\rm bGPT_{libri}$ on LibriSpeech for audio or $\rm bGPT_{image}$ on ImageNet for images) are fine-tuned on tasks of another modality, showing non-trivial improvements over random initialization. This suggests shared byte patterns between modalities like audio and images. However, negative transfer is observed when transitioning between text and other modalities, indicating that the benefits of structured pattern learning in pre-training do not universally apply. Text, as a human-created abstraction, exhibits distinct byte-level organizational patterns that differ significantly from audio and visual data, which may explain the negative transfer effects observed in text-related cross-modal fine-tuning.

To further investigate cross-modal knowledge transfer in byte models, we evaluated their performance on the Speech Commands v2 dataset, transformed into 32$\times$32 BMP spectrograms. This process, converting 8KB audio files to 3KB images, inevitably leads to information loss. However, our focus shifts towards exploring the nuances of knowledge transfer dynamics rather than mere performance gains. We selected two pre-trained models for evaluation: $\rm bGPT_{image}$ for its data format consistency with spectrograms and $\rm bGPT_{libri}$ for its information similarity with spectrograms (both involving speech). These models were employed for conditional generative modelling and classification, following a procedure akin to that in Table 3.

\begin{table}[t]
\centering
\caption{Cross-modal knowledge transfer efficacy of bGPT models on spectrogram-based Speech Commands v2.}
\vspace{1em}
\begin{tabular}{lcc}
\toprule
\textbf{Model} & \textbf{BPB} & \textbf{Acc (\%)} \\
\midrule
$\rm bGPT_{random}$ & 0.4052 & 80.74 \\
$\rm bGPT_{image}$  & \textbf{0.3899} & 84.66 \\
$\rm bGPT_{libri}$  & 0.3900   & \textbf{85.26} \\
\bottomrule
\end{tabular}
\vspace{-1em}
\end{table}

The results of Table 4 reveal closely matched BPB between $\rm bGPT_{image}$ and $\rm bGPT_{libri}$, suggesting that the disparity observed in their CIFAR-10 performance does not extend to this spectrogram-based task. This is because CIFAR-10, comprising natural scenes, shares fewer patterns with spectrograms compared to the commonalities between spectrograms and raw audio. Notably, $\rm bGPT_{libri}$, pre-trained on audio, achieves higher accuracy than $\rm bGPT_{image}$ on spectrograms with speech content, indicating effective cross-modal transfer and the importance of content alignment over mere data format. This suggests that byte models possess an inherent capability to discern and translate abstract data features and patterns that are independent of the specific format or modality of the input.

Together, these results confirm the versatility and adaptability of bGPT models in processing digital media data and transferring knowledge across modalities, even when data formats differ. This underscores the potential of byte models for integrating shared knowledge from various sources to enhance digital world comprehension.

\subsection{Algorithm and Hardware Simulation}
The primary objective of this evaluation was to assess the capabilities of bGPT in simulating algorithms and hardware. Due to the lack of baseline models and widely used datasets, we investigated its performance across different data scales, evaluating the scalability of bGPT on binary data. This was carried out through two tasks: data conversion and CPU state modelling, utilizing data scales ranging from $\rm 10^3$ ($\mathrm{bGPT^3}$), $\rm 10^4$ ($\mathrm{bGPT^4}$), $\rm 10^5$ ($\mathrm{bGPT^5}$), to $\rm 10^6$ ($\mathrm{bGPT^6}$). To ensure that the results were primarily driven by data scale, all models were randomly initialized. For data conversion, we utilized the IrishMAN dataset \cite{DBLP:conf/hcmir/WuLY023}, which includes paired ABC notation and MIDI files, evaluated on scales from $\rm 10^3$ to $\rm 10^5$. For CPU state modelling, we used the CPU States dataset, evaluated on scales from $\rm 10^4$ to $\rm 10^6$.

\begin{figure}[t]
    \centering
    \begin{minipage}[t]{0.45\textwidth}
        \centering
        \includegraphics[width=0.95\linewidth]{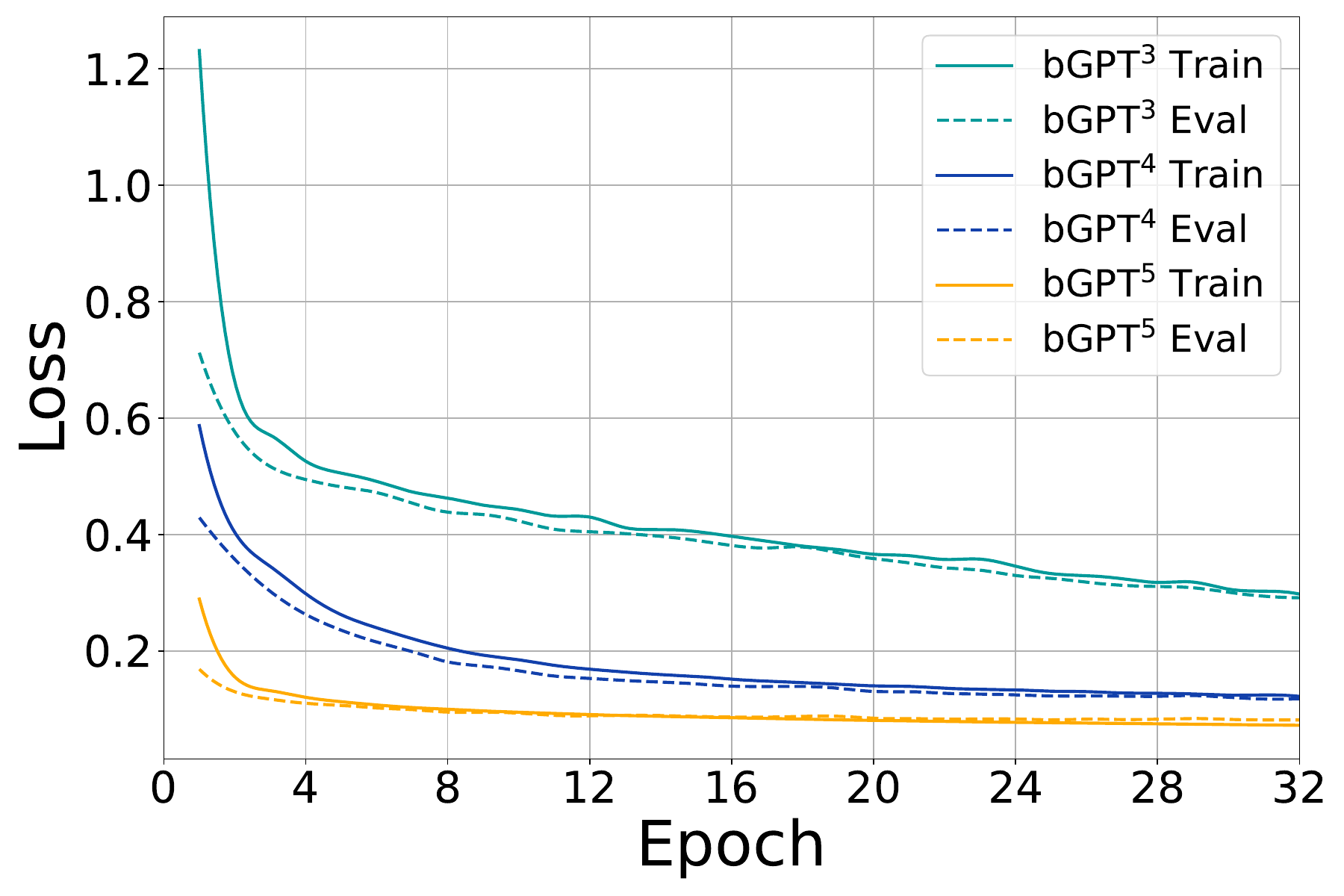}
        \\ (a) Data Conversion
        \vspace{1em}
    \end{minipage}
    \vspace{1em}
    \begin{minipage}[t]{0.45\textwidth}
        \centering
        \includegraphics[width=0.95\linewidth]{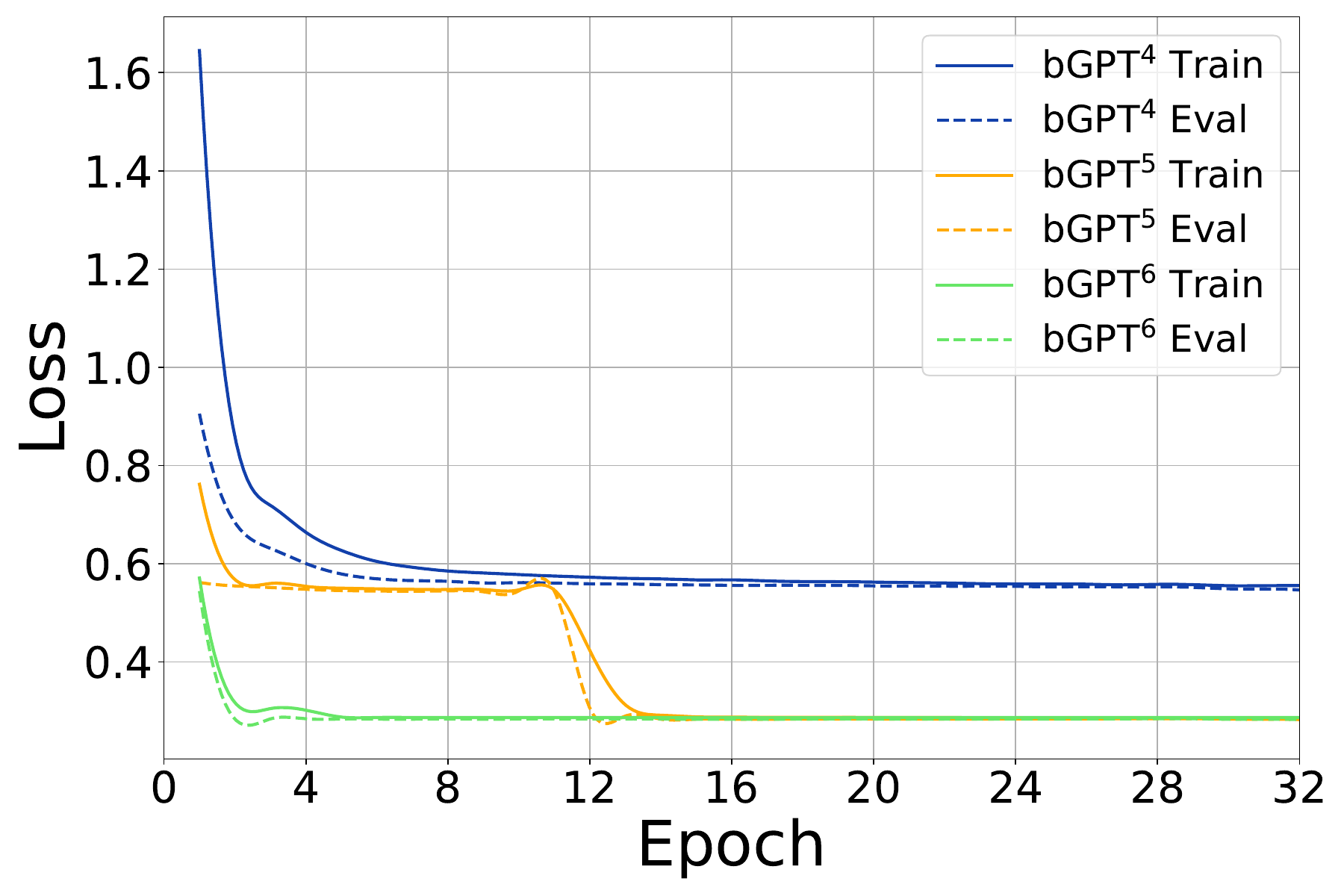}
        \\ (b) CPU State Modelling
    \end{minipage}
    \vspace{-1em}
    \caption{Training and evaluation loss curves for different data scales in bGPT performance across epochs.}
    \vspace{-0.5em}
\end{figure}

\begin{table}[t!]
\centering
\caption{Performance comparison of bGPT across different data scales in data conversion.}
\vspace{1em}
\resizebox{\linewidth}{!}{
\begin{tabular}{lcccc}
\toprule
& \multicolumn{2}{c}{\textbf{ABC $\rightarrow$ MIDI}} & \multicolumn{2}{c}{\textbf{MIDI $\rightarrow$ ABC}} \\
\cmidrule(lr){2-3} \cmidrule(lr){4-5}
\textbf{Model} & \textbf{$\rm \mathbf{BPB_{abc}}$} & \textbf{$\rm \mathbf{BPB_{midi}}$} & \textbf{$\rm \mathbf{BPB_{midi}}$} & \textbf{$\rm \mathbf{BPB_{abc}}$} \\
\midrule
$\mathrm{bGPT^3}$ & 1.4319 & 0.2381 & 0.2796 & 1.1227 \\
$\mathrm{bGPT^4}$      & 0.9789 & 0.0202 & 0.1751 & 0.1763 \\
$\mathrm{bGPT^5}$     & 0.7362 & 0.0011 & 0.1341 & 0.0789 \\
\bottomrule
\end{tabular}
}
\vspace{-0.5em}
\end{table}

\begin{table}[t!]
\centering
\caption{Performance comparison of bGPT across different data scales in CPU state modelling.}
\vspace{1em}
\begin{tabular}{lcc}
\toprule
\textbf{Model} & \textbf{BPB} & \textbf{Acc (\%)} \\
\midrule
$\mathrm{bGPT^4}$ & 0.6924 & 27.2404 \\
$\mathrm{bGPT^5}$  & 0.0969 & 99.9702 \\
$\mathrm{bGPT^6}$  & 0.0969   & 99.9961 \\
\bottomrule
\end{tabular}
\vspace{-0.5em}
\end{table}

\subsubsection{Data Conversion}
The task of converting between ABC notation and MIDI evaluates the capability of byte models to simulate algorithms that work with both human-interpretable (ABC notation) and native binary (MIDI) data formats. While previous studies avoided direct MIDI modelling due to its binary nature \cite{DBLP:conf/iclr/HuangVUSHSDHDE19,DBLP:conf/acl/ZengTWJQL21,DBLP:conf/aaai/HsiaoLYY21}, byte models are naturally suited for this task, as they can directly work with binary data.

We separately calculated BPB values for each format to clearly measure model performance. For example, in ABC to MIDI conversion, $\rm BPB_{abc}$ assesses generative modelling, as the model generates content from scratch. While $\rm BPB_{midi}$ evaluates data conversion, considering the complete ABC byte sequence is given.

We observe a lower starting loss and a more rapid convergence in Fig. 3a as the data scale grows, indicating that increased data volume directly enhances model performance in simulating the data conversion process. From Table 5, we see that as the scale of data increases, BPB for both ABC to MIDI and MIDI to ABC conversions decrease significantly. The $\mathrm{bGPT^5}$ model achieves an impressively low $\rm BPB_{midi}$ of 0.0011 in ABC to MIDI conversion, which is extremely close to perfect performance (where BPB reaches 0), surpassing the performance of the $\mathrm{bGPT^3}$ model by orders of magnitude.

Table 5 suggests consistently higher BPB for ABC in both directions, which is likely due to two factors: 1) The forward conversion from ABC to MIDI focuses on simulating an existing algorithm with necessary information, while the reverse process from MIDI to ABC requires inferring and reconstructing missing information in MIDI files like score structures, musical ornaments, and expressions. 2) As MIDI is a lower-level binary format and ABC notation is a human-readable text format, byte models may find it easier to learn patterns within MIDI files.

\subsubsection{CPU State Modelling}
CPU state modelling aims to replicate CPU functionality by predicting updates to internal states from machine instructions. We employed bGPT for predicting these states, selecting the highest probability byte at each step based on complete instructions and initial states. The accuracy was assessed through byte-wise comparisons with actual states.

We discovered that data volume significantly influences modelling performance. Table 6 shows significant performance variations, with a notable BPB drop from $\mathrm{bGPT^4}$ to $\mathrm{bGPT^5}$, but diminishing returns beyond $\mathrm{bGPT^5}$. Both $\mathrm{bGPT^5}$ and $\mathrm{bGPT^6}$ achieved near-perfect accuracies (99.97\% and 99.99\%), suggesting an efficiency beyond simple memorization, given that each test case contained an average of 128 random instructions, and the vast potential combinations of instruction scenarios (over 516 million).

A significant improvement in the performance of $\mathrm{bGPT^5}$ occurred around epoch 11, as shown in Fig. 3b, indicating an emergent ability in CPU state modelling. This leap, especially in BPB and accuracy when comparing $\mathrm{bGPT^4}$ and $\mathrm{bGPT^5}$, suggests a deeper understanding of CPU states may stem from a qualitative enhancement in capability. This aligns with the concept of emergent abilities in large LMs \cite{DBLP:journals/tmlr/WeiTBRZBYBZMCHVLDF22}, where capabilities seem to spontaneously arise with scale and complexity.

However, scepticism exists regarding whether these improvements reflect genuine learning \cite{schaeffer2023are}. Critics argue the performance boosts might be due to non-linear metrics or overfitting. Nonetheless, the linear and smooth nature of BPB counters this, indicating that the improvements likely stem from a real comprehension of CPU operations, suggesting consistent learning rather than metric anomalies.

In summary, bGPT demonstrates strong scalability on native binary data with emergent abilities in data conversion and CPU state modelling, which illuminate its potent capabilities in algorithm and hardware simulation tasks. While the tasks utilized for demonstration in this study were not excessively complicated, the near-perfect performance observed in these contexts points to the broader potential of byte models for simulating and reverse-engineering a wide range of algorithms and hardware.

\section{Conclusions}
In this paper, we present bGPT as a versatile simulator for the digital world, extending deep learning to binary data processing via next byte prediction. Our experiments demonstrate the effectiveness of bGPT in modelling digital media data, which showcases modality-agnostic knowledge transfer. We observe a strong scalability of bGPT in modelling native binary data and even signs of emergent abilities. bGPT performs comparably to specialized models across diverse datasets without modality-specific designs, and excels in data conversion and CPU state modelling, demonstrating its potential for simulating various algorithms and hardware.

Nonetheless, our experiments illuminate opportunities for improvement. In this study, we confine the modelling to short audio segments and low-resolution images, a consequence of the resource-intensive nature intrinsic to byte models. Due to limited computational resources, we only investigated data conversion between ABC notation and MIDI, without broader assessments across alternate formats. Furthermore, to simplify data collection and evaluation, our CPU state modelling experiments focused solely on simplified CPUs, omitting the use of real modern CPUs, which are considerably more complex.

Future research directions for byte models include: 1) reducing computational cost to make training byte models more feasible; 2) scaling models and dataset sizes to accommodate a broader range of native binary data, as well as handling larger digital media files such as high-resolution images and videos; and 3) improving model performance, particularly for underexplored tasks involving native binary data across diverse application domains.

\section{Impact Statements}
In this paper, we introduce bGPT, a model designed for binary data processing and digital world modelling, pushing the boundaries of deep learning into the realm of native binary data. This innovation enables bGPT to directly interpret and manipulate binary data, offering profound insights into digital systems. While bGPT presents advancements in understanding and modelling the digital world, it also necessitates a careful examination of its ethical implications and potential impact on societal norms and legal frameworks.

Its ability to simulate or reverse-engineer algorithms and hardware has two major implications: 1) it can significantly boost technological innovation, aiding in the development of cybersecurity, software, and hardware by understanding and improving on existing technologies; 2) it poses a risk to intellectual property, as training bGPT on extensive datasets of paired source code and executable software, might enable the reverse-engineering of proprietary software. This capability, while showcasing its potential, could facilitate unauthorized access to or modification of software, raising security and legal issues.

In conclusion, while bGPT and similar byte models offer exciting opportunities for advancing our understanding and capabilities within the digital world, their deployment requires thoughtful consideration of the ethical, societal, and legal implications. This is especially crucial for safeguarding intellectual property and ensuring security against potential malicious use.

\bibliography{example_paper}
\bibliographystyle{icml2024}

\newpage
\appendix
\onecolumn

\begin{figure}[t!] 
    \centering
    \includegraphics[width=\textwidth]{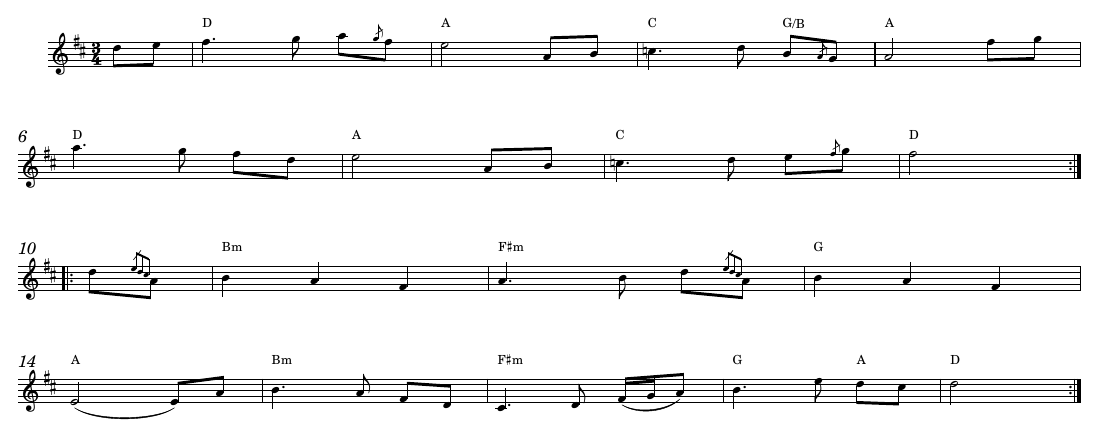} 
    \caption{Sheet music example corresponding to the ABC notation of a folk tune in D major.}
\end{figure}

\section{Symbolic Music Formats}
\subsection{ABC Notation}
ABC notation is a shorthand form of music notation that uses ASCII characters to represent musical notes and symbols. It is a compact, human-readable, and easily typed format that has gained popularity for its ability to facilitate the swift digital transcription, sharing, and editing of music. Originating within folk and traditional music circles, ABC notation has become a universal method for musicians and enthusiasts to exchange music without the need for proprietary software or extensive musical training.

To illustrate the fundamentals of ABC notation, consider the example below, which depicts a folk tune:

\begin{lstlisting}[style=abcnotation]
X:1
L:1/8
M:3/4
K:D
de |"D" f3 g a{/g}f |"A" e4 AB |"C" =c3 d"G/B" B{/A}G |"A" A4 fg |
"D" a3 g fd |"A" e4 AB |"C" =c3 d e{/f}g |"D" f4 ::
d{/edc}A |"Bm" B2 A2 F2 |"F#m" A3 B d{/edc}A |"G" B2 A2 F2 |
"A" (E4 E)A |"Bm" B3 A FD |"F#m" C3 D (F/G/A) |"G" B3 e"A" dc |"D" d4 :|
\end{lstlisting}

The corresponding sheet music rendered from this ABC sequence is shown in Fig. 4. In this example, the structure of ABC notation starts with a header that includes:

\begin{itemize}
\item \textbf{X (Tune Number)}: This is a unique number assigned to the tune, serving as a reference in collections or databases.
\item \textbf{L (Unit Note Length)}: This defines the default duration of notes. In this context, \texttt{1/8} indicates that each note is an eighth note, which is a standard choice for folk tunes.
\item \textbf{M (Meter)}: This denotes the time signature of the piece, which determines the rhythmic pulse. Here, \texttt{3/4} suggests a waltz-like rhythm.
\item \textbf{K (Key)}: This sets the key signature for the tune. \texttt{D} signifies that the piece is in the key of D major.
\end{itemize}

After the header, the body of the ABC notation consists of the musical notes themselves. Each note is represented by a letter (\texttt{A} to \texttt{G}), which corresponds to the pitch. The octaves are indicated by lower-cased letters or apostrophes for higher octaves and upper-cased letters or commas for lower or higher octaves, respectively. The duration of notes is modified by numbers; for instance, a \texttt{2} following a note would double its value, while a \texttt{/2} would halve it. Chords can be notated by enclosing the chord name in quotation marks to suggest harmony, such as \texttt{"F\#m"} representing the F sharp minor. Additional symbols are used for sharps \texttt{\^}, flats \texttt{\_}, and naturals \texttt{=} to alter pitch, and various other characters represent articulations and expressions.

\begin{figure}[t!] 
    \centering
    \includegraphics[width=\textwidth]{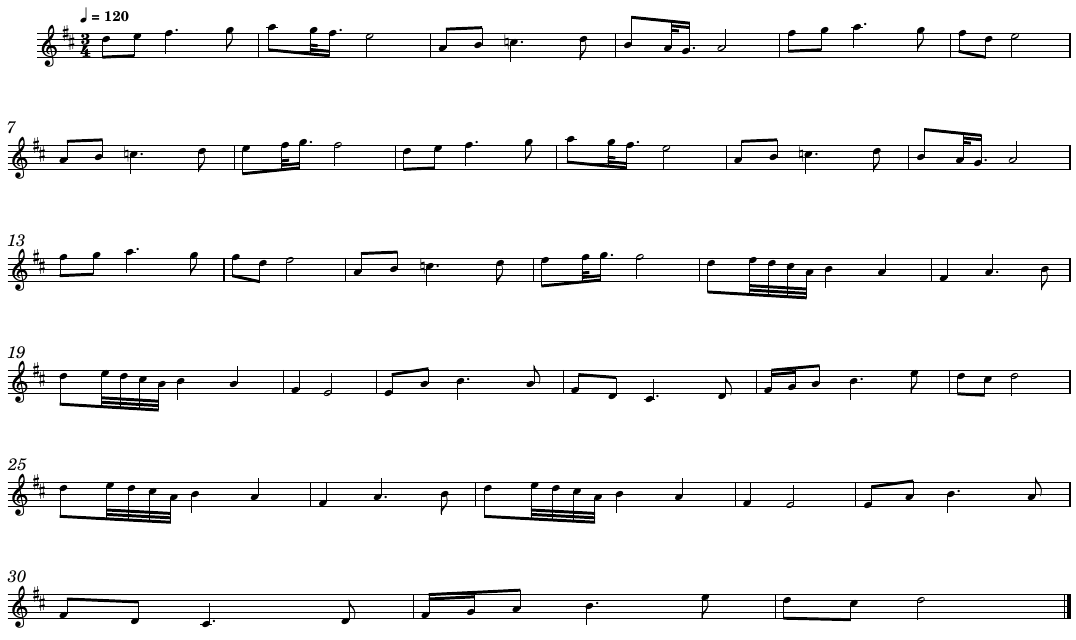} 
    \caption{Sheet music example corresponding to the MIDI of the same folk tune presented in Fig. 4.}
\end{figure}

This notation describes the tune with sufficient information, from its rhythmic structure down to the details of its melody and harmony. As such, ABC notation serves as a bridge between the traditional form of musical scores and the modern digital world, offering a text-based alternative that is both efficient and widely accessible. For a detailed documentation of ABC notation, please refer to the ABC notation standard\footnote{\url{https://abcnotation.com/wiki/abc:standard:v2.1}}.

\subsection{MIDI}
MIDI, short for Musical Instrument Digital Interface, is a technical standard that describes a protocol, digital interface, and connectors and allows a wide variety of electronic musical instruments, computers, and other related devices to connect and communicate with one another. Unlike ABC notation, which is designed for human readability and the transcription of musical scores, MIDI is focused on capturing and conveying information about music performance and production in a format that machines can interpret and process.

The core of MIDI communication involves MIDI messages, which are digital signals that represent various musical elements such as notes, pitch, velocity (how hard a note is played), control changes (modifications to sound parameters), and program changes (instrument changes). These messages do not transmit audio; instead, they send instructions about how music should be played or produced. This approach allows MIDI to be incredibly versatile and lightweight compared to raw audio, suitable for real-time control during live performances and detailed editing in studio environments.

MIDI files, commonly with the extension .mid or .midi, store sequences of MIDI messages that can be played back by computer software or electronic instruments. These files make it possible to share and distribute musical compositions in a form that can be easily modified, transposed, or played back with different instrument sounds, offering a level of flexibility and creative freedom that is unique to the digital music domain.

There are several ways to create and edit MIDI, each catering to different workflows and preferences:

\begin{itemize}
\item \textbf{MIDI Keyboards and Controllers}: Physical devices that allow musicians to input notes and control data into a computer. They emulate traditional instruments and can capture the nuances of a performance in MIDI format.

\item \textbf{Piano Roll}: A graphical interface in digital audio workstations (as shown in Fig. 1) where notes are displayed and edited on a grid representing pitch and time. It is intuitive for composing melodies, chords, and rhythms.

\item \textbf{Score Editor}: Allows users to input and edit music in traditional notation, converting written notes into MIDI data. Ideal for those familiar with musical notation.

\item \textbf{Programming and Scripting}: Involves writing code to generate and manipulate MIDI data, offering complex and generative musical possibilities beyond manual input methods. For instance, complex MIDI sequences and automation can be generated using software such as Max/MSP, Pure Data, or programming languages (e.g., SuperCollider).
\end{itemize}

Although MIDI offers immense benefits for musicians and producers, when converting MIDI data back into score-oriented symbolic music formats like ABC notation, certain challenges arise due to the intrinsic differences between these formats. MIDI excels at capturing how a piece of music is played, including the timing, velocity, and expressiveness of each note. In contrast, ABC notation focuses on the structural elements of music, such as melody, harmony, and rhythm, using a textual format that is more accessible for human reading and writing but less detailed in terms of performance nuances.

To illustrate the conversion between MIDI and ABC notation, consider the following example:

\begin{lstlisting}[style=abcnotation]
X:1
L:1/8
M:3/4
K:D
de f3 g | ag/<f/ e4 | AB =c3 d | BA/<G/ A4 | fg a3 g | fd e4 |
AB =c3 d | ef/<g/ f4 | de f3 g | ag/<f/ e4 | AB =c3 d | BA/<G/ A4 |
fg a3 g | fd e4 | AB =c3 d | ef/<g/ f4 | de/4d/4c/4A/4 B2 A2 | F2 A3 B |
de/4d/4c/4A/4 B2 A2 | F2 E4 | EA B3 A | FD C3 D | F/G/A B3 e | dc d4 |
de/4d/4c/4A/4 B2 A2 | F2 A3 B | de/4d/4c/4A/4 B2 A2 | F2 E4 | EA B3 A |
FD C3 D | F/G/A B3 e | dc d4 |]
\end{lstlisting}

This is the same tune we have demonstrated in Fig, 4, but it has now been converted from ABC notation to MIDI and subsequently back to ABC notation. The corresponding sheet music, rendered from this ABC sequence, is shown in Fig. 5. One can observe that the ABC sequence and the musical score appear significantly longer than the original due to the elimination of repeat signs in MIDI, requiring repeated sections to be explicitly written out. Additionally, the bar divisions and the ornaments in the MIDI-to-ABC conversion might not align perfectly with the original ABC notation. These ornaments are often represented as a series of rapid notes in MIDI, which can complicate their translation back into ABC notation.

While MIDI-rendered sheet music and its ABC notation counterpart may sound identical when performed, the visual complexity of the MIDI transcription can make it more challenging to read than the ABC version. This discrepancy underscores the fact that MIDI and ABC are formats tailored for different contexts. MIDI is adept at capturing nuanced performance data, such as dynamics and subtle timing variations, which are essential for playback fidelity but are not as visually straightforward as ABC notation. On the other hand, ABC notation, with its simplified symbolic representation, is not ideal for conveying detailed performance information.

Both formats have overlapping capabilities in representing musical information, yet each also possesses unique characteristics that do not translate perfectly when converting from one to the other. This results in a loss of information: nuances specific to each format can be diminished or omitted entirely during the conversion process. Such losses present a challenge in preserving the full intent and detail of a musical piece when moving between ABC and MIDI formats.


\section{CPU States Dataset}
The CPU States Dataset has been crafted using a Python script to simulate the operations of a CPU, offering a controlled environment to assess the capabilities of bGPT in modelling the hardware behaviour. This appendix offers an overview of the dataset, introducing the structure, instruction set, and details of the dataset curation.

The dataset comprises entries of 1KB memory blocks, each paired with a subsequent 16-byte sequence representing the CPU register states. The memory blocks encode varying numbers of machine instructions, while the register state sequence includes the CPU status post-execution of each instruction. The CPU registers in each state consist of:

\begin{itemize}
\item \textbf{Program Counter (PC)}: 1 byte in size, points to the next instruction in the memory.
\item \textbf{Accumulator (ACC)}: A 1-byte register involved in arithmetic and logic operations.
\item \textbf{Instruction Register (IR)}: Occupies 4 bytes, storing the currently executed instruction.
\item \textbf{General-Purpose Registers}: An additional 10 bytes are allocated, emulating a range of registers (labelled A-J) used for various computational purposes.
\end{itemize}

Transitions between states mimic the execution cycle of a CPU, including fetching instructions from memory, decoding them to determine the required operation, and executing the operation to modify the state. The execution process includes instructions for modifying register values, performing arithmetic calculations, and controlling the program flow.

Each instruction in the dataset is encoded in a four-byte format \texttt{OP ADDR1 ADDR2 ADDR3}, with each byte serving a specific purpose:

\begin{itemize}
\item \textbf{OP (Operation Code)}: 1 byte representing the opcode of the instruction to be executed.
\item \textbf{ADDR1, ADDR2, ADDR3}: 1 byte each, representing up to three addresses that the instruction may interact with.
\end{itemize}

The opcode and addresses together describe the action to be taken by the CPU. The instructions vary in terms of the number of addresses they use, from zero-address to three-address instructions. This variability allows the simulation of a wide range of operations, from simple data movement to complex arithmetic and logical functions.

Table 7 (located on page 19) details the instruction set and the variants each instruction can have, depending on the values in the address fields. The variant to be executed is determined by the non-zero value in the address fields. For zero-address instructions, all address fields are zero. For single-address instructions, ADDR1 holds the address of the register involved, and the other fields are zero. For two- and three-address instructions, the respective fields hold the addresses of the registers involved in the operation. 

In designing the instruction set, complex instructions like jump commands were deliberately excluded to prevent the occurrence of infinite loops or other complexities that could hinder straightforward simulation.

The Python script generates the CPU States Dataset by randomly initializing the ACC and general-purpose registers for each simulation, thereby ensuring a wide range of starting conditions. It then executes a random sequence of instructions, varying in length from 1 to 256, with the effect of each instruction on the CPU state being recorded. This results in a sequence of states that reflects the cumulative impact of the executed instructions. The script guarantees that each sequence culminates with a halt instruction \texttt{HLT}, marking the end of a program. Furthermore, when considering only the register changes related to instructions, there are over 516 million potential scenarios resulting from combinations of various instruction types, registers, and their values. 

The dataset thus produced comprises 2.1 million entries for the training set and 21,000 entries for the evaluation set. This process, with its random initial states and varied instruction lengths, produces a diverse dataset that is essential for training and evaluating the effectiveness of bGPT in modelling CPU states.

To aid in a more intuitive comprehension of the operational logic of the CPU, a randomly generated program for executing 10 instructions along with their corresponding states in human-readable form is provided below:

\begin{lstlisting}[style=cpu]
Program: ['MUL J A', 'DIV I', 'MUL E D', 'ADD C', 'LOADI 86', 'MOV A H', 'AND D E', 'POP H', 'CLR', 'HLT']

State at step 0:
PC: 0
ACC: 146
IR: HLT
Registers: {'A': 19, 'B': 55, 'C': 245, 'D': 35, 'E': 174, 'F': 185, 'G': 9, 'H': 20, 'I': 140, 'J': 2}

State at step 1:
PC: 1
ACC: 146
IR: MUL J A
Registers: {'A': 19, 'B': 55, 'C': 245, 'D': 35, 'E': 174, 'F': 185, 'G': 9, 'H': 20, 'I': 140, 'J': 2}

State at step 2:
PC: 2
ACC: 146
IR: DIV I
Registers: {'A': 19, 'B': 55, 'C': 245, 'D': 35, 'E': 174, 'F': 185, 'G': 9, 'H': 20, 'I': 140, 'J': 38}

State at step 3:
PC: 3
ACC: 1
IR: MUL E D
Registers: {'A': 19, 'B': 55, 'C': 245, 'D': 35, 'E': 174, 'F': 185, 'G': 9, 'H': 20, 'I': 140, 'J': 38}

State at step 4:
PC: 4
ACC: 1
IR: ADD C
Registers: {'A': 19, 'B': 55, 'C': 245, 'D': 35, 'E': 255, 'F': 185, 'G': 9, 'H': 20, 'I': 140, 'J': 38}

State at step 5:
PC: 5
ACC: 246
IR: LOADI 86
Registers: {'A': 19, 'B': 55, 'C': 245, 'D': 35, 'E': 255, 'F': 185, 'G': 9, 'H': 20, 'I': 140, 'J': 38}

State at step 6:
PC: 6
ACC: 86
IR: MOV A H
Registers: {'A': 19, 'B': 55, 'C': 245, 'D': 35, 'E': 255, 'F': 185, 'G': 9, 'H': 20, 'I': 140, 'J': 38}

State at step 7:
PC: 7
ACC: 86
IR: AND D E
Registers: {'A': 20, 'B': 55, 'C': 245, 'D': 35, 'E': 255, 'F': 185, 'G': 9, 'H': 20, 'I': 140, 'J': 38}

State at step 8:
PC: 8
ACC: 86
IR: POP H
Registers: {'A': 20, 'B': 55, 'C': 245, 'D': 35, 'E': 255, 'F': 185, 'G': 9, 'H': 20, 'I': 140, 'J': 38}

State at step 9:
PC: 9
ACC: 86
IR: CLR
Registers: {'A': 20, 'B': 55, 'C': 245, 'D': 35, 'E': 255, 'F': 185, 'G': 9, 'H': 86, 'I': 140, 'J': 38}

State at step 10:
PC: 10
ACC: 0
IR: HLT
Registers: {'A': 20, 'B': 55, 'C': 245, 'D': 35, 'E': 255, 'F': 185, 'G': 9, 'H': 86, 'I': 140, 'J': 38}
\end{lstlisting}

The dataset is intended for use in machine learning research, particularly for developing models capable of predicting CPU behaviour. It can also serve as an educational tool, helping students understand CPU operation principles.

\begin{table}[t]
\centering
\caption{Defined instruction set and the address variants of different instructions.}
\vspace{1em}
\begin{tabular}{|l|l|p{10cm}|}
\hline
\textbf{Instruction} & \textbf{Address Variants} & \textbf{Meaning} \\
\hline
HLT & 0 & Halts the CPU with ``HLT" \\
\hline
CLR & 0, 1 & Zeroes the accumulator with ``CLR" or a specific register with ``CLR A" \\
\hline
INC & 0, 1 & Increments the accumulator with ``INC" or a specific register with ``INC A" \\
\hline
DEC & 0, 1 & Decrements the accumulator with ``DEC" or a specific register with ``DEC A" \\
\hline
SHL & 0, 1 & Shifts the accumulator left by one bit with ``SHL" or a specific register with ``SHL A" \\
\hline
SHR & 0, 1 & Shifts the accumulator right by one bit with ``SHR" or a specific register with ``SHR A" \\
\hline
ROL & 0, 1 & Rotates the accumulator left by one bit with ``ROL" or a specific register with ``ROL A" \\
\hline
ROR & 0, 1 & Rotates the accumulator right by one bit with ``ROR" or a specific register with ``ROR A" \\
\hline
NOT & 0, 1 & Performs bitwise NOT on the accumulator with ``NOT" or a specific register with ``NOT A" \\
\hline
PUSH & 1 & Pushes the value of a specific register onto the stack with ``PUSH A" \\
\hline
POP & 1 & Pops the top of the stack into a specific register with ``POP A" \\
\hline
LOADI & 1 & Loads an immediate value into the accumulator with ``LOADI value" \\
\hline
SWAP & 1, 2 & Swaps values between the accumulator and a register with ``SWAP A" or between two registers with ``SWAP A B" \\
\hline
ADD & 1, 2, 3 & Adds using one, two, or three addresses with examples ``ADD A", ``ADD A B", ``ADD A B C" \\
\hline
SUB & 1, 2, 3 & Subtracts using one, two, or three addresses with examples ``SUB A", ``SUB A B", ``SUB A B C" \\
\hline
MUL & 1, 2, 3 & Multiplies using one, two, or three addresses with examples ``MUL A", ``MUL A B", ``MUL A B C" \\
\hline
DIV & 1, 2, 3 & Divides using one, two, or three addresses with examples ``DIV A", ``DIV A B", ``DIV A B C" \\
\hline
AND & 1, 2, 3 & Performs bitwise AND using one, two, or three addresses with examples ``AND A", ``AND A B", ``AND A B C" \\
\hline
OR & 1, 2, 3 & Performs bitwise OR using one, two, or three addresses with examples ``OR A", ``OR A B", ``OR A B C" \\
\hline
XOR & 1, 2, 3 & Performs bitwise XOR using one, two, or three addresses with examples ``XOR A", ``XOR A B", ``XOR A B C" \\
\hline
MOV & 2 & Moves a value from one register to another with ``MOV A B" \\
\hline
\end{tabular}
\label{table:instruction_set}
\end{table}

\end{document}